\documentclass{llncs}

\usepackage{graphicx}
\usepackage{amsmath}
\usepackage{multirow}
\usepackage{amsfonts}
\usepackage{amsmath}
\usepackage{mathtools}
\usepackage{amssymb}

\begin{document}

\title{A machine learning based heuristic to predict the efficacy of online sale}
\author{Aditya Vikram Singhania$^{1,*}$, Saronyo Lal Mukherjee$^1$, Ritajit Majumdar$^2$, Akash Mehta$^1$, Priyanka Banerjee$^1$ \and Debasmita Bhoumik$^{2,+}$}
\institute{$^1$Department of Computer Science,
The Bhawanipur Education Society College\\
$^2$Advanced Computing \& Microelectronics Unit, Indian Statistical Institute, India\\
\email{$^{*}$mailingadityavs@gmail.com,
$^{+}$debasmita.ria21@gmail.com}}

\maketitle

\begin{abstract}
It is difficult to decide upon the efficacy of an online sale simply from the discount offered on commodities. Different features have different influence on the price of a product which must be taken into consideration when determining the significance of a discount. In this paper we have proposed a machine learning based heuristic to quantify the \textit{``significance"} of the discount offered on any commodity. Our proposed technique can quantify the significance of the discount based on features and the original price, and hence can guide a buyer during a sale season by predicting the efficacy of the sale. We have applied this technique on the Flipkart Summer Sale dataset using Support Vector Machine, which predicts the efficacy of the sale with an accuracy of 91.11\%. Our result shows that very few mobile phones have a significant discount during the Flipkart Summer Sale.

\end{abstract}

\section{Introduction}

Online retailers such as Amazon, Flipkart etc. provide lucrative sale offers in almost every season. Discount offered for various products often lures the customers into buying items which may not be of actual or immediate requirement to them. However, a question which can hardly be answered from these sale offers is whether the discount offered for a particular product is \textit{significant} for its features. For example, a laptop with a nominal SSD storage can be priced higher than another laptop with a higher HDD storage. The same discount for the former is arguably better than that for the later. Therefore, it is necessary to take into consideration the influence of individual features towards the price of the commodity, and then decide on the significance of the discount provided. A machine learning based technique is proposed in this paper to decide on the efficacy of online sale which can guide a buyer by predicting the \textit{``significance"} of the discount offered on any commodity.

Machine learning (ML) is the technique of training a model to perform a specific task without using explicit instructions \cite{alpaydin2020introduction}. ML algorithms can either be initially trained with some data so that it can perform predictions over similar, but previously unseen dataset; or it can be expected to extract patterns from the dataset all by itself. The former is termed as Supervised Learning \cite{caruana2006empirical}, while the later is the Unsupervised Learning \cite{ghahramani2003unsupervised}. Apart from these, other learning algorithms such as Reinforcement learning \cite{kaelbling1996reinforcement}, Semi-supervised learning \cite{zhu2009introduction} etc. also have a wide range of application.

Fields such as air fare prediction \cite{tziridis2017airfare}, stock market analysis \cite{nazario2017literature}, real-estate price prediction \cite{park2015using} etc. have seen a plenty of applications of ML. In this paper, we have proposed a ML based heuristic to predict the efficacy of the sale price for any commodity. Our technique partitions the commodities into \textit{discount classes} which essentially determines the \textit{amount of} \textit{significance} of a discount. In other words, we quantify the \textit{``significance"} of the discount offered for a commodity, as a function of the standard deviation of its price class, in order to guide the buyer to decide whether he/she should buy it or not.

The proposed technique is then applied on the \textit{Flipkart summer sale} to decide the \textit{``significance"} of the discount offered for mobile phones during this sale. To obtain the results for the said sale, we have (i) created our own sale and non-sale dataset through web crawling; (ii) heuristically classified the set of mobile phones into four price classes and have used Support Vector Machine to train the model with the data of non-sale season; (iii) applied the trained model on the sale dataset to predict the price class of each mobile phone based on their features with 91.11\% accuracy; (iv) determined the \textit{``significance"} of the discount for each mobile phone. Our results show that Flipkart offered a \textit{significant} discount for very few mobile phones. Although this result is largely dependent on the dataset and the heuristic grouping, our proposed technique is universal and can be applied on any dataset with any number of groups created through any heuristic.

The rest of the paper is organized as follows: In Section 2 we discuss the proposed mathematical technique for quantifying \textit{``significance"} of a discount price. Section 3 elaborates the Web Crawling technique used to create the dataset on which we have applied the proposed technique. Sections 4 and 5 respectively presents the training of ML algorithm for the created dataset and the resultant efficacy as obtained by our technique. We conclude in Section 6.

\section{Quantifying the efficacy of online sale}
Consider an E-commerce website which sells $N$ commodities (of the same type, e.g. mobile phones or laptops), where the original (non-sale) and discounted (sale) price of the $i^{th}$ commodity are $n_i$ and $s_i$ respectively. The set of $N$ commodities are first partitioned into $g$ price classes according to their price range. Let $G_j$, $1 \leq j \leq g$ denote the $j^{th}$ price class. Each price class $G_j$ is associated with a price range $[p_j^{start},p_j^{end}]$, and a commodity $i$ is assigned to $G_j$ only if its \emph{non-sale} price $n_i \in [p_j^{start},p_j^{end}]$. Therefore, $\sum_{j=1}^{g} |G_j| = N$, and $G_k \cap G_l = \phi$, $\forall$ $1 \leq k \neq l \leq g$. For each price class $G_j$, we also determine the mean price $\Bar{G_j}$ and the standard deviation (S.D.) $\sigma_{G_j}$. Indeed, $\sigma_{G_j}$ governs the variation of price for the commodities in the price class $G_j$. Therefore, we have considered the \textit{level of significance} of a discount as the number of standard deviations which the discount price is away from the original price.

For each commodity, a set of features $\textbf{f} = \{f_1, \hdots, f_m\}$ are selected, and an ML algorithm is trained, where each training data is a tuple ($\textbf{f}_l, G_l$), with $\textbf{f}_l$ as the feature vector for the $l^{th}$ commodity and $G_l$ as its assigned price class.

In the sale dataset, for each commodity $i$, we run the ML algorithm to predict its assigned price class $G_i$. Furthermore, let $G_{d_i}$ be the price class for the discount price for the same commodity. Our heuristic is that:
\begin{enumerate}
    \item If $G_{d_i}$ is lower than $G_{i}$, i.e., the discount takes the commodity to a lower price class, then we define the discount to be \textbf{`$\infty$-fold significant'}.
    \item If $G_{d_i} = G_i$, then the discount is defined to be \textbf{`n-fold significant'} if $(n+1)k\sigma_{G_i} \leq s_i - d_i < nk\sigma_{G_i}$, for some fixed $k \in \mathbb{R}$ and $0 < k \leq 1$; $n \in \mathbb{Z}^{+}$.
\end{enumerate}

This quantification of \textit{``significance"}, as presented above, allows a lot of flexibility. In our original definition of \textbf{`$\infty$-fold significant'} we did not consider how much lower $G_{d_i}$ is from $G_{i}$. However, if necessary, one can define \textbf{`$\infty_x$-fold significant'} if $G_{d_i}$ is $x$ classes lower than $G_{i}$. Obviously \textbf{`$\infty_x$-fold significant'} is better than \textbf{`$\infty_{x-h}$-fold significant'} $\forall$ $h \geq 1$. Furthermore, since $k \in \mathbb{R}$ and $n \in \mathbb{Z}^{+}$, it is possible to make $k$ arbitrarily close to 0, and $n$ arbitrarily large. However, making $k$ very small will lead to a large number of \textit{`folds'} with very small range for each and hence most of such ranges will most likely remain empty. Making $n$ large will soon result in moving from one price class to another, and then it falls in the realm of \textbf{`$\infty$-fold significant'}. One interesting scope of study, which we have not looked at in this paper, is how best to define the significance class for a given dataset.

The flexibility in quantifying \textit{``significance"} and the use of ML to determine the influence of different features on the price makes our technique usable in a wide range of application. Apart from the prediction of sale efficacy, this technique can as well be used for more complex tasks, such as, one can learn the choice of the folds from prior knowledge of the buying pattern of a person, and use our technique as a recommender system for that buyer during a sale.

In this paper, we have applied this technique for determining the efficacy of the discount provided in \textit{Flipkart Summer Sale} for mobile phones.

\section{Creation of the dataset}
There are quite a few datasets available for mobile price prediction in the internet. However, we could not find any dataset in which the price of phones, both during a sale and the original price, were available. Therefore we created our own dataset via web crawling.

A web crawler is an internet bot that systematically browses the internet for the purpose of web indexing. It starts with a list of URLs to visit, called the seeds. As the crawler visits each URL, it identifies all the hyperlinks in that page and adds them to the list of URLs to visit henceforth. The web crawler can be used to copy and save the information of each page that it visits. A similar technique, called web data extraction, is deployed to use HTML tags in order to identify important information from the visited URL, and the extracted data, according to those tags, is saved in a local database. For our dataset creation, we have used two Python library modules, namely (i) \emph{request}, which sends HTTP requests to the URL and returns a Response Object with all the response data (content, encoding, status, etc), and (ii) \emph{beautifulsoup4} \cite{richardson2007beautiful}, which is used for data extraction from HTML, XML and other markup languages.

For the purpose of this research, we have used data extraction on the Flipkart website only for mobile phones. The first sale datatset was created via web data extraction on $19^{th}$ March 2020, when the \emph{Flipkart Summer Sale} was active, and the second non-sale dataset was created on $24^{th}$ March, 2020 after the termination of the summer sale. The features which we have considered in this paper are (i) RAM, (ii) Storage, (iii) Megapixel of Front Camera, (iv) Megapixel of Back Camera, (v) Battery Capacity, and (vi) Internet connectivity. Our crawling resulted in a dataset with 1193 entries. However, we have excluded all those basic phones which do not have internet connection, and have henceforth worked with the first five features only. This is because those basic phones had a significant deviation in the other five features from the smart phones, which could lead to a biased training of the ML algorithm. Furthermore, hardly any people buy those phones now, and it can be a safe assumption that not many will be interested to buy the basic phones at discounted rates.

The crawled dataset resulted in some entries where one or more of the features had null values. After dropping all such data, and also dropping duplicate data (same mobile set sold in different colors), we were left with 733 entries, which constitutes our final dataset. The non-sale and sale price of each mobile phone, along with the features, were stored separately for the training and testing purposes respectively.

\subsection{Heuristic grouping of the price classes}

For the prediction of sale efficacy with this dataset, we partitioned the crawled dataset into four price classes as depicted in Table~\ref{tab:class}. For this particular choice of price classes, we also calculate the mean ($\mu$) and standard deviation ($\sigma$) of each price class.

\begin{table}[htb]
    \centering
    \caption{Heuristic grouping of the price classes}
    \begin{tabular}{|c|c|c|c|c|}
    \hline
       \textbf{Price Range} & \textbf{Assigned class} & \textbf{\# phones} & \textbf{$\mu$} & \textbf{$\sigma$} \\
       \hline
        (0, 5000) & ``LOW" & 426 & 1350.92 & 1051.63\\
        \hline
        [5000, 15000) & ``BUDGET" & 204 & 9476.41 & 3007.14\\
        \hline
        [15000, 30000) & ``MID RANGE" & 76 & 19483.81 & 3914.78\\
        \hline
        $>$ 30000 & ``PREMIUM" & 27 & 51372.44 & 19097.43\\
        \hline
    \end{tabular}
    \label{tab:class}
\end{table}

% \begin{figure}[htb]
%     \centering
%     \caption{Number of phones in each price class}
%     \includegraphics[scale=0.5]{Main Project/mobiles.png}
%     \label{fig:class}
% \end{figure}

We note from Table~\ref{tab:class} that the grouping into classes is not symmetric, and the distribution of the number of phones is skewed towards the lower price classes. This heuristic for the grouping was developed via an informal survey carried out by the authors amongst their friends and relatives, where they were asked to define the price range for which they will mark a mobile phone as `low cost', `budget phone', `mid range' or `premium'. From that survey, this grouping seemed to be the most obvious one. However, one can readily use the original technique (from Section 2) to form a different set of price classes.

\section{Training the model with Support Vector Machine}

We now use the proposed technique from Section 2 on our crawled dataset to predict the efficacy of discount on mobile phones during \textit{Flipkart summer sale} with four price classes. For the quantification of \textit{``significance"}, we have considered the fixed value of $k = \frac{1}{2}$. Furthermore, we have truncated the number of \textbf{`n-fold significant'} discount classes to three based on the result obtained from our dataset. Increasing the number of such classes would have only resulted in multiple classes with zero or very few entries. The quantification of the \textbf{`n-fold significant'} classes, and their assigned names are depicted in Table~\ref{tab:discount}.

\begin{table}[htb]
    \centering
    \caption{Quantification and nomenclature of \textit{``significance"} classes}
    \begin{tabular}{|c|c|}
    \hline
        \textbf{Range of discount} & \textbf{Assigned \textit{``significance"} class name} \\
        \hline
        $0 \leq s_i - d_i < \frac{\sigma}{2}$ & ``POOR"\\
        \hline
        $\frac{\sigma}{2} \leq s_i - d_i < \sigma$ & ``ACCEPTABLE"\\
        \hline
        $s_i - d_i > \sigma$ & ``GOOD" \\
        \hline
        Different price class & ``EXCELLENT"\\
        \hline
    \end{tabular}
    \label{tab:discount}
\end{table}

In other words, the \textbf{`$\infty$-fold significant'} class has been renamed as ``EXCELLENT", and by definition, it is not possible to have ``EXCELLENT" deals for ``LOW" price class.

Machine Learning is now used to train the model with the non-sale dataset to predict the price class of any mobile phone based on the features mentioned earlier. We have used Support Vector Machine (SVM) Algorithm for the training purpose.

\subsection{SVM Algorithm and training with non-sale data}

\begin{figure}[htb]
    \centering
    \caption{Linear Support Vector Machine}
    \includegraphics[scale=2.5]{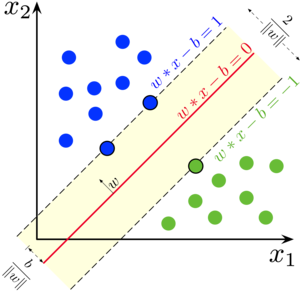}
    \label{fig:svm1}
\end{figure}

Classification \cite{kotsiantis2007supervised} is a subclass of Supervised Learning. A $k$-class classification problem, where $C_1, \hdots, C_k$ are the class labels, consists of $m$ training samples and $n$ testing samples. Each training sample is a tuple $(s_l,C_{s_l})$ where the later is the designated class of $s_l$. After the training phase, for each test data $t_l$, the algorithm predicts a class $C_i$ such that
$$Prob(t_l \in C_i) > Prob(t_l \in C_j), \forall j \neq i.$$ 
Support Vector Machine (SVM) \cite{cortes1995support} is one of the most extensively used supervised machine learning algorithm for classification problem \cite{nurhanim2017classification,dey2019automatic}. In SVM, the input is a set of labeled training data and the output is an optimal hyper-plane. The $m$ training sets are of the form $(x_1 , y_1 ), (x_2 , y_2 ), . . . (x_m , y_m )$, where $x_i \in \mathbb{R}^d$ is a feature, and the class label is $y_i \in \{+1, -1\}$, $i = 1 ... m$. SVM creates an optimal separating hyper-plane based on a mathematical function known as the kernel function. A kernel function is defined as $k(X_i,X_j) = \phi (X_i^T) \phi(X_j)$, where $X_i,X_j$ are feature vectors and $\phi$ is the kernel function. Fig.~\ref{fig:svm1} shows a the hyper-plane created by SVM for a linearly separable dataset using linear kernel function.
% one side of the  The data points with feature vectors belong to one side of the hyper-plane is in the negative class, and the others are in the positive class.
% The working mechanism of linear support vector machine is shown in Fig \ref{fig:svm1}

% \\

However, every dataset does not contain vectors which are linearly separable. In such cases, we require a nonlinear SVM classifier which uses other forms of kernel function such as polynomial kernel, RBF kernel and sigmoid kernel.

In the heuristic grouping, elaborated in Section 3, there are four price classes, and hence this is a multi-class classification problem. There are two techniques \cite{bishop2006pattern} for multi-class SVM: (i) One-against-one which combines several binary classifiers, and (ii) One-against-all which considers all data at once.

The entire non-sale dataset was divided into the training and testing part where 80\% of the dataset was used for training purpose and the remaining 20\% for testing purpose. Training was done using the SVM algorithm with the RBF kernel and the One-against-all technique, as they were predicted best for our data via the cross-validation \cite{hastie2009elements} method. The training and testing accuracy thus obtained were 98.66\% and 91.11\% respectively.

\section{\textit{``Significance"} of discount in Flipkart Summer Sale}

We have applied the trained SVM model on the Flipkart Summer Sale mobile phone dataset. The price classes and the \textit{``significance"} class are as depicted in Table~\ref{tab:class} and ~\ref{tab:discount} respectively. The SVM algorithm predicts an overall 710 ``POOR", 7 ``ACCEPTABLE", 3 ``GOOD", and 13 ``EXCELLENT" discounts. We have selected two mobile phones from our dataset to show the detailed calculation of the \textit{``significance"} of discount for them, where one is an ``ACCEPTABLE" discount and the other one is an ``EXCELLENT" discount. The summary of our result is depicted in Table~\ref{tab:result}.

The feature vector \textbf{f} for each of the two phones follow the ordering: (RAM, Storage, Front camera, Back camera, Battery capacity), and the price vector \textbf{p} follows the ordering: (original price, sale price). \textbf{o} and \textbf{s} denotes the price class for original and sale price respectively, and \textbf{$\sigma_o$} is the S.D. of the original price class.

\begin{enumerate}
    \item \textbf{Company}: iBall, \textbf{f}: (2, 16, 0, 5, 2800), \textbf{p}: (4399, 3749), \textbf{o}: Low, \textbf{s}: Low, \textbf{$\sigma_o$}: 1051.63. Since $\frac{\sigma_o}{2} \leq s_i - d_i = 650 < \sigma_o$, this discount is ``ACCEPTABLE".
    
    \item \textbf{Company}: LG, \textbf{f}: (4, 128, 5, 16, 3300), \textbf{p}: (60000, 14999), \textbf{o}: Premium, \textbf{s}: Budget. Since the discount price falls in a lower price class, this discount is ``EXCELLENT".
\end{enumerate}

\begin{table}[ht!]
    \centering
    \caption{\textit{``Significance"} of discount in Flipkart Summer Sale}
    
    \begin{tabular}{|c|c|c|c|c|c|}
         \hline
        Discount class name & \multicolumn{5}{|c|}{\textit{``significance"} of discount} \\
         \hline
          & Excellent & Good & Acceptable & Poor & Total \\
         \hline
         Low & 0 & 1 & 1 & 424 & 426 \\
         \hline
         Budget & 4 & 1 & 3 & 196 & 204 \\
         \hline
         Mid-range & 5 & 1 & 2 & 68 & 76 \\
         \hline
         Premium & 4 & 0 & 1 & 22 & 27 \\
         \hline
         \hline
         Total & 13 & 3 & 7 & 710 & 733 \\
         \hline
    \end{tabular}
    \label{tab:result}
\end{table}

From Table~\ref{tab:result} it can be readily inferred that Flipkart does not provide significant discount on most of the mobile phones during its Summer Sale season.

\section{Conclusions and Future Work}

In this paper, we have proposed a ML based technique to quantify the \textit{``significance"} of the discount offered by online retailers for a commodity during sale season. Our proposed technique quantifies the \textit{``significance"} of discount based on the features, original price and the discount class of the product. The application of this technique on Flipkart Summer Sale shows with an accuracy of 91.11\% that significant discount is offered on very few mobile phones during this sale. We intend to extend the application part of this paper for other commodities, other sale seasons, and other online retailers as well. Such a cross-platform comparative study can be used as a guiding system for a buyer during online sale.

\bibliographystyle{unsrt}
\bibliography{main}

\end{document}